\documentclass[10pt,twocolumn,letterpaper]{article}

\usepackage[pagenumbers]{cvpr} 

\usepackage[dvipsnames]{xcolor}

\definecolor{cvprblue}{rgb}{0.21,0.49,0.74}
\usepackage[pagebackref,breaklinks,colorlinks,citecolor=cvprblue]{hyperref}
\usepackage{algorithm}
\usepackage{algpseudocode}
\usepackage{amsmath}
\usepackage{graphicx}
\usepackage{makecell}
\usepackage{booktabs}
\usepackage{caption}
\usepackage{colortbl}
\usepackage{xcolor}
\usepackage{multirow}
\usepackage{soul}
\usepackage{wrapfig}
\newcommand{\tablestyle}[2]{\setlength{\tabcolsep}{#1}\renewcommand{\arraystretch}{#2}\centering\small}
\newcommand{\snowflake}{\includegraphics[width=0.32cm,height=0.32cm]{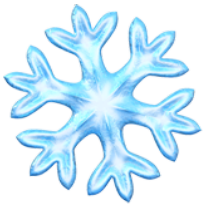}}
\newcommand{\fire}{\includegraphics[width=0.32cm,height=0.32cm]{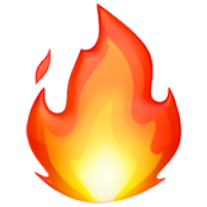}}
\def\paperID{} 
\def\confName{3DV\xspace}
\def\confYear{2025\xspace}
\newcommand{\mname}{DreamBeast}%

\title{
\begin{minipage}{\textwidth}
    \centering
    \includegraphics[trim={1cm 1.2cm 1.2cm 1.3cm},clip,width=1cm]{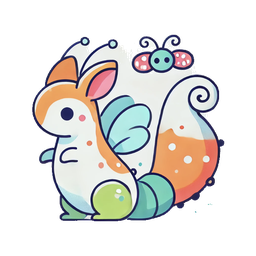} 
    DreamBeast: Distilling 3D Fantastical Animals with Part-Aware Knowledge Transfer
    \vspace{-1em}
\end{minipage}
}

\author{%
  Runjia Li$^1$\:
  \quad
  Junlin Han$^1$\:
  \quad
  Luke Melas-Kyriazi$^1$\:
  \quad
    Chunyi Sun$^2$\:
  \quad
    Zhaochong An$^3$\:\\
  \quad
    Zhongrui Gui$^1$\:
  \quad
  Shuyang Sun$^1$\:
  \quad
  Philip Torr$^1$\:
  \quad
  Tomas Jakab$^1$\:
\\
  $^1$University of Oxford
  \quad
  $^2$Australian National University
   \quad
  $^3$University of Copenhagen
  \\
   {\small \url{dreambeast3d.github.io}}
}
\begin{document}
\twocolumn[\maketitle\vspace{-3em} \begin{center}
    \includegraphics[width=0.92\textwidth,height=250px]{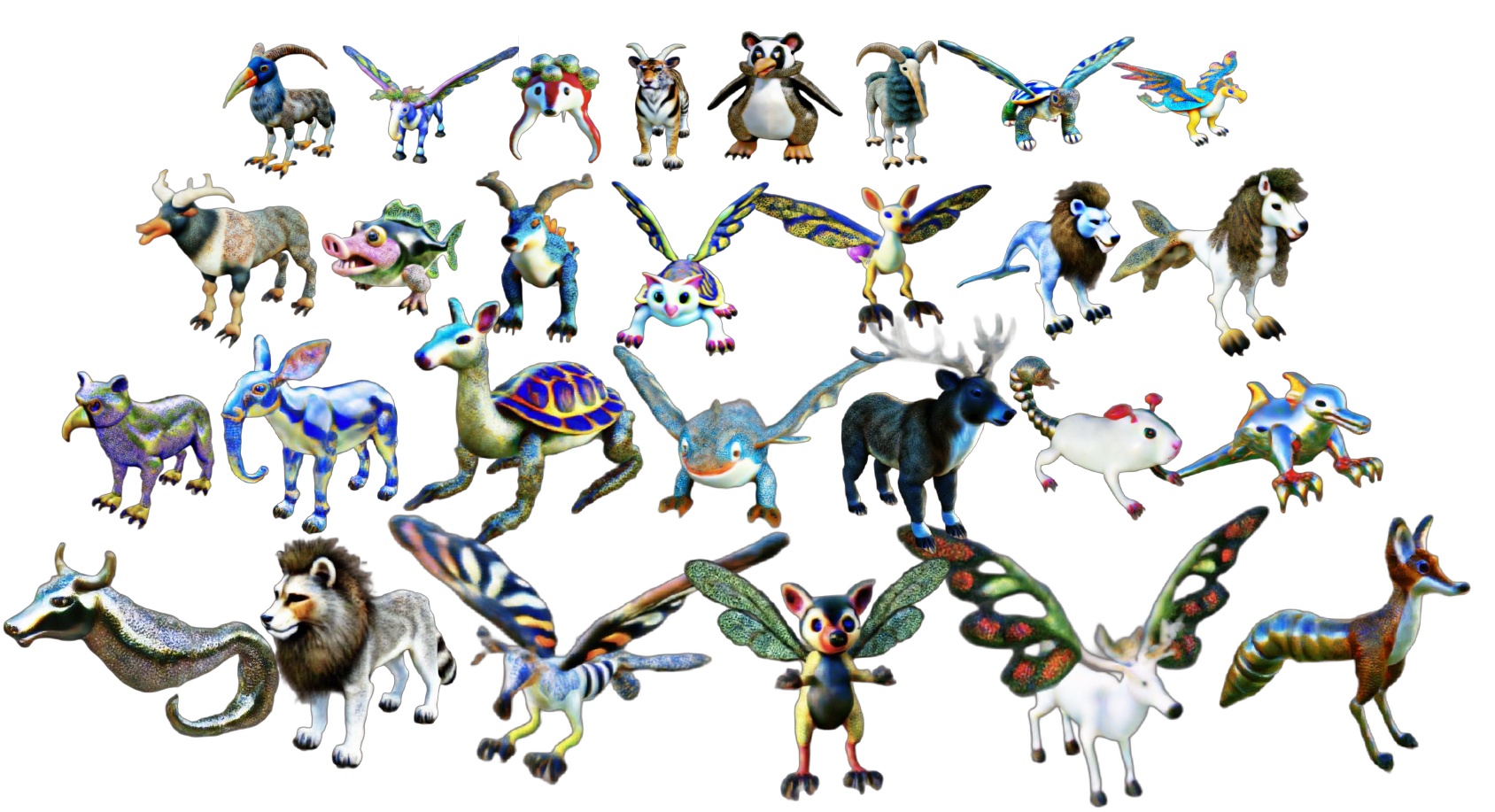}
\end{center}
\captionof{figure}{
\textbf{Generated fantastic 3D beasts composed of diverse animal parts.} Our method enables part-level generation, resulting in 3D creatures with unique combinations of heads, limbs, wings, tails, and bodies. 
}\bigbreak]

\begin{abstract}
We present \mname, a novel method based on score distillation sampling (SDS) for generating fantastical 3D animal assets composed of distinct parts. 
Existing SDS methods often struggle with this generation task due to a limited understanding of part-level semantics in text-to-image diffusion models. 
While recent diffusion models, such as Stable Diffusion 3, demonstrate a better part-level understanding, they are prohibitively slow and exhibit other common problems associated with single-view diffusion models. 
\mname~overcomes this limitation through a novel part-aware knowledge transfer mechanism. 
For each generated asset, we efficiently extract part-level knowledge from the Stable Diffusion 3 model into a 3D Part-Affinity implicit representation. 
This enables us to instantly generate Part-Affinity maps from arbitrary camera views, which we then use to modulate the guidance of a multi-view diffusion model during SDS to create 3D assets of fantastical animals. 
\mname~significantly enhances the quality of generated 3D creatures with user-specified part compositions while reducing computational overhead, as demonstrated by extensive quantitative and qualitative evaluations.
\end{abstract}

\section{Introduction}
Imagine a creature taking flight, its dragon wings catching the sunlight. Picture its majestic lion's head surveying the landscape, while a sinuous serpent's tail trails behind.  What if "Fantastic Beasts and Where to Find Them" was not just a magical story, but we could actually build them in a digital 3D world?
Current methods for generating 3D objects~\cite{hong2023lrm, han2024vfusion3d, wu2016learning, chen2019text2shape, song2019generative, cai2020learning, zhou20213d} struggle with generating complex, artistic, or fantastical shapes and textures, which are not represented in existing datasets. For example, they are unable to produce Griffin-like animals composed of parts from multiple species.
More generally, they struggle with producing objects composed of multiple diverse parts.

One of the most promising current approaches to open-world 3D asset generation consists of lifting 2D guidance into 3D.
Methods such as DreamFusion~\cite{dreamfusion} and SJC~\cite{sjc} demonstrate how pre-trained 2D diffusion models~\cite{deepfloyd, sd21, sd3, mvdream} can guide the generation of 3D objects through score distillation sampling (SDS). 
Specifically, these methods produce 3D objects from textual descriptions by utilizing the priors encoded in 2D diffusion models, which act as approximate log gradients of the density of distribution of 2D images conditioned on text.

The lifting methods, however, fall short of providing part-level controllability for part-specific textual descriptions. 
The reason for this is twofold. 
First, there have not been any 2D diffusion models capable of sufficiently strong part-level understanding. 
Second, in part due to the first reason, there have been no methods proposed in the literature for part-aware lifting-based (SDS) text-to-3D generation from part-specific textual prompts.

\begin{figure}[h!]
    \centering
    \includegraphics[width=0.47\textwidth]{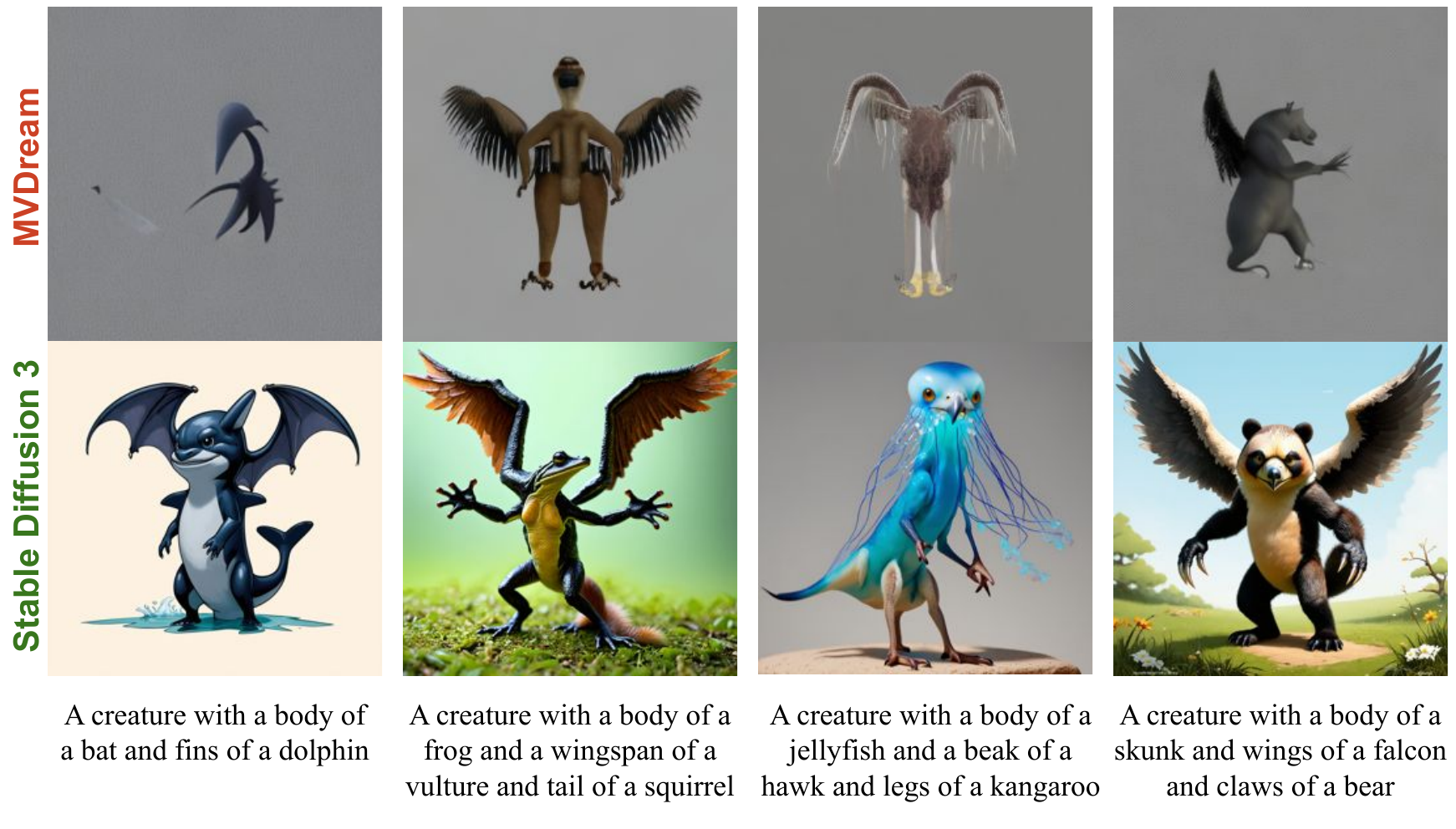}
    \caption{\textbf{Comparison of diffusion models on part-level prompt understanding in 2D generation.} Although MVDream can grasp the overall semantic understanding of the described animals, the generated images often feature deformed animals and fail to accurately capture specific part-based descriptions, unlike SD3.}
    \label{fig:diffusion_comparison}
\end{figure}
\begin{figure}[h!]
    \centering
    \includegraphics[width=0.38\textwidth]{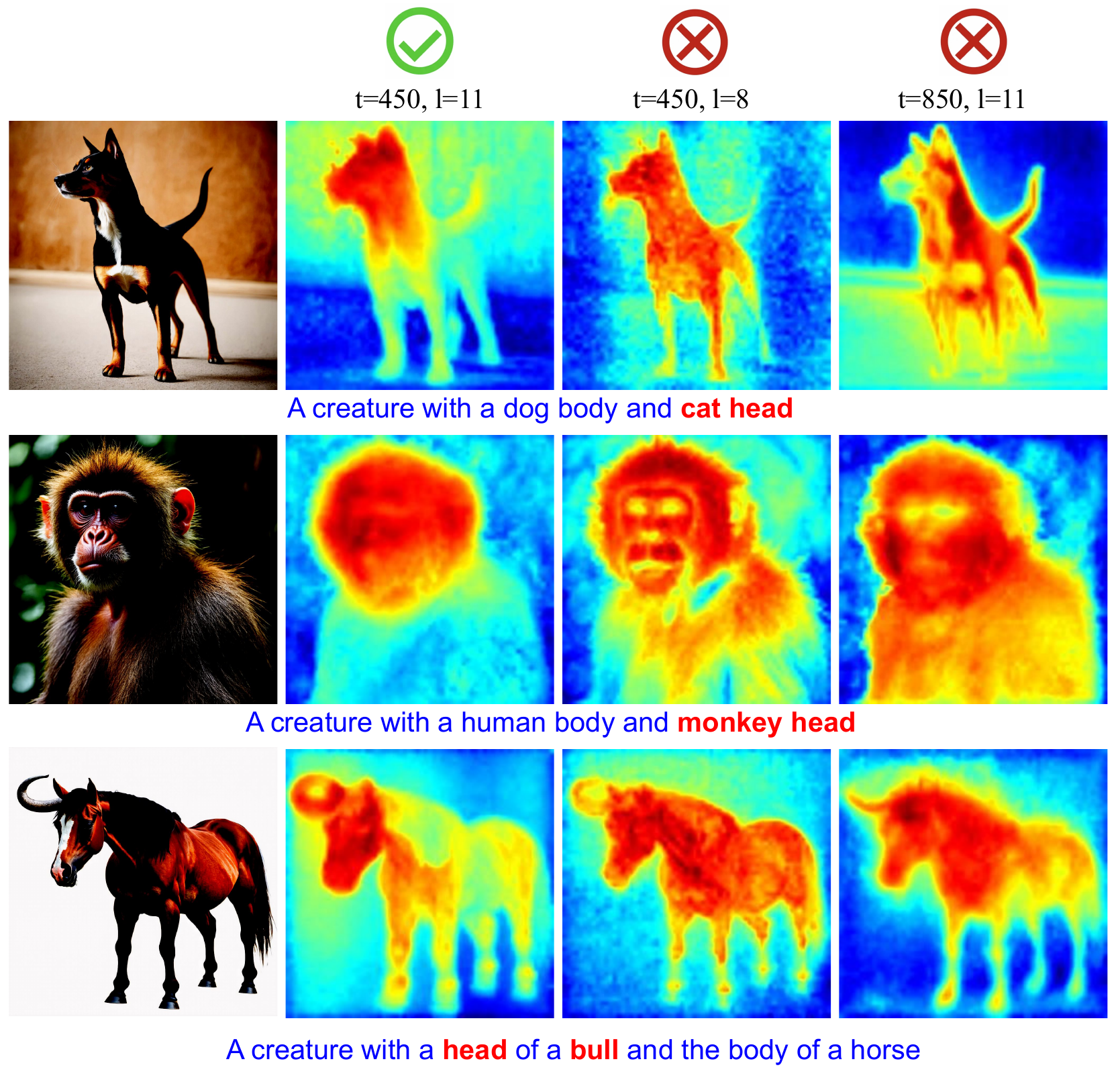}
    \caption{\textbf{Failing to generate part-aware content even with part understanding in SD3. }
    Despite its understanding of part correspondences, as evidenced by the cross-attention maps at certain timesteps $t$ and layers $l$, SD3 may still fail to generate part-aware images. This is illustrated in above examples where specific animal parts are absent, highlighted in red.
    Our method capitalizes on the observation that only particular timesteps $t$ and layers $l$ exhibit part-awareness.}
    \label{fig:sd3_cross_attn}
\end{figure}
\begin{figure}[h!]
    \centering
    \includegraphics[width=0.42\textwidth]{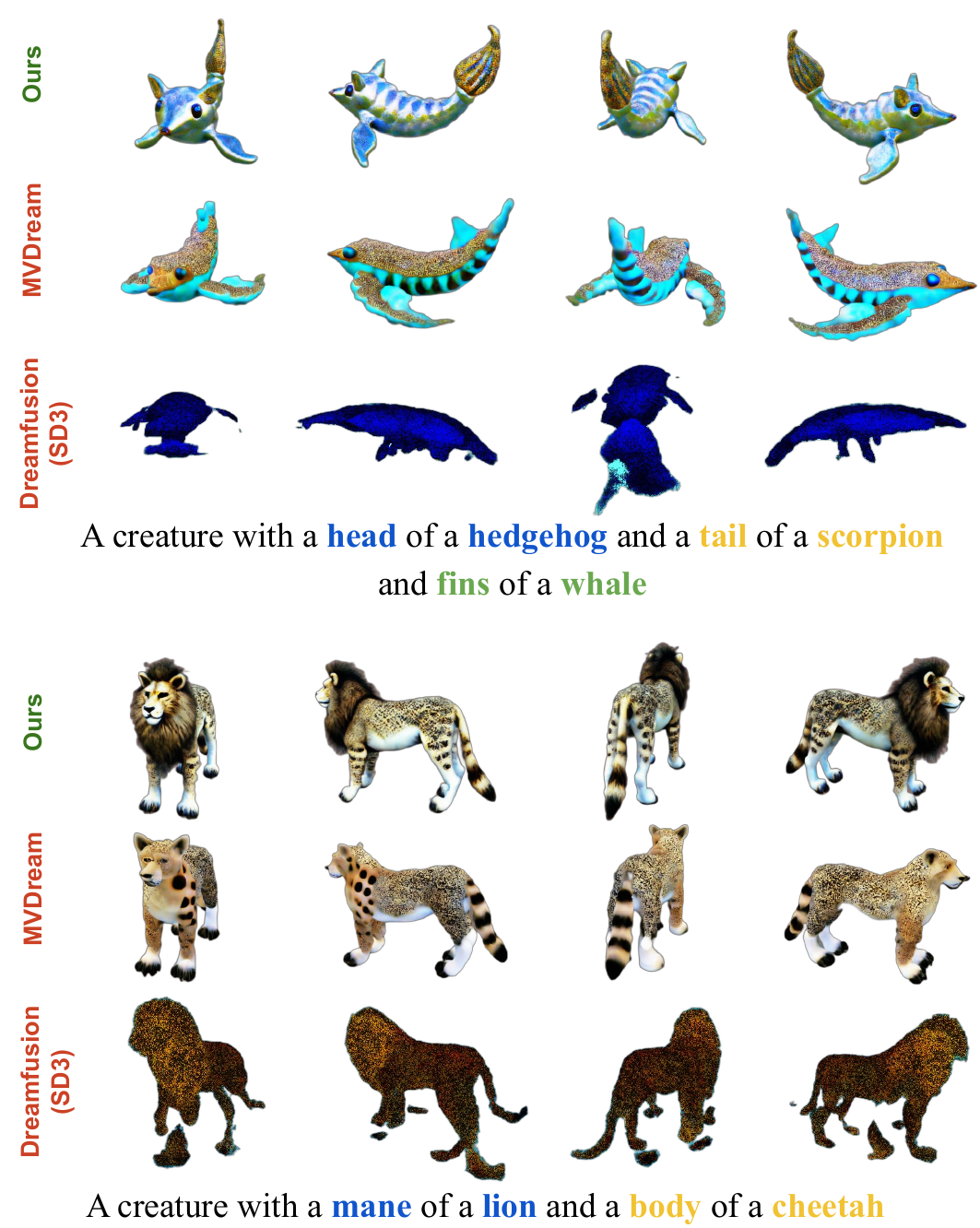}
    \caption{\textbf{MVDream and SD3 have difficulty generating part-aware 3D animals.} While SD3~\cite{sd3} can understand part correspondence in images and text, it struggles to generate 3D assets using SDS due to the issues we discussed in our paper. MVDream~\cite{mvdream} falls short because it was fine-tuned on Objaverse~\cite{objaverse}, which lacks part-level information in the dataset.}
    \label{fig:intro_demo}
\end{figure}
\begin{figure}[h!]
    \centering
    \includegraphics[width=0.35\textwidth]{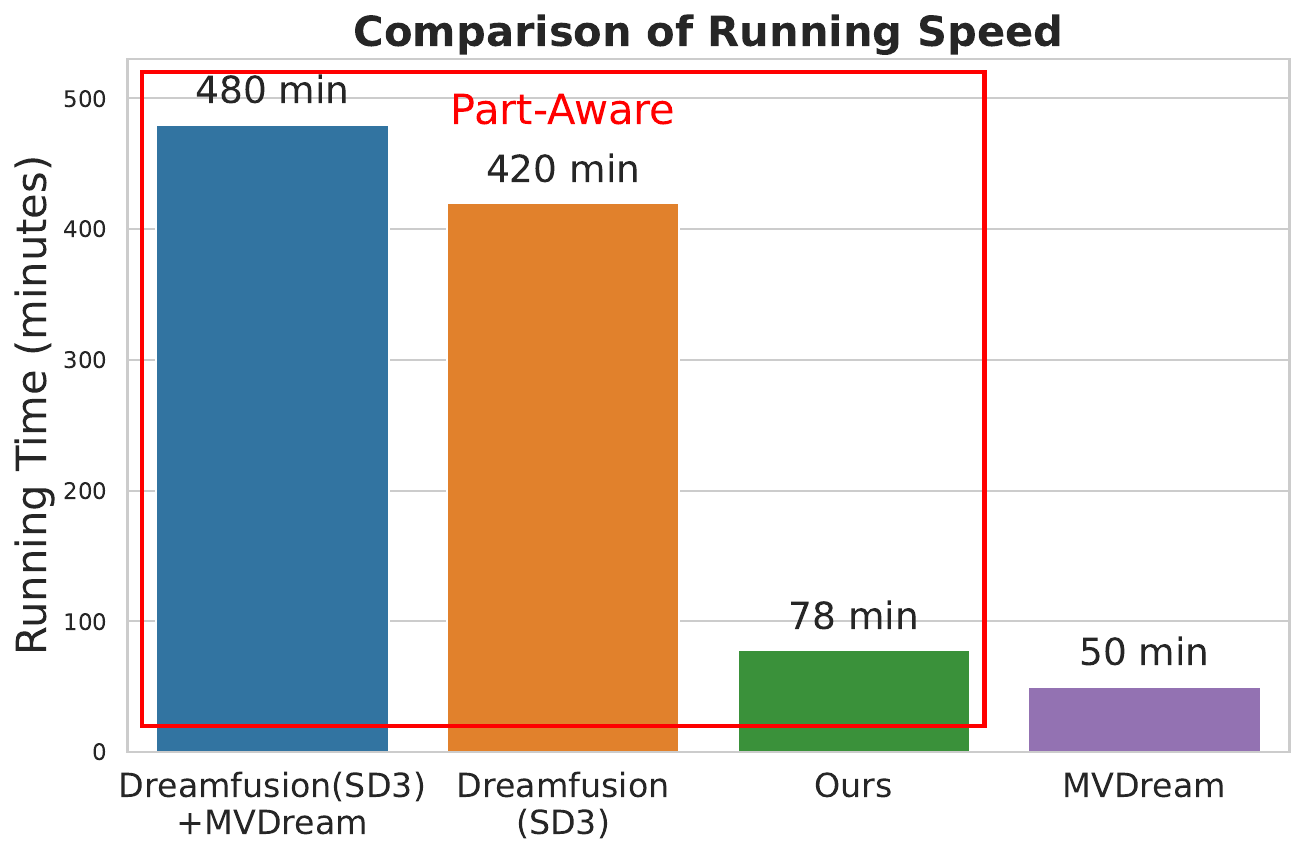}
    \caption{\textbf{Running speed comparison.} While Dreamfusion (SD3) combined with MVDream and standalone Dreamfusion (SD3) take 480 and 420 minutes respectively, our method significantly reduces the runtime to 78 minutes. This reduction is achieved without sacrificing part-awareness making our method both faster and more effective in part-aware 3D generation. More detail in Appendix~\ref{as:speed}.}
    \label{fig:running_speed}
\end{figure}

Diffusion models are improving rapidly, and the recent release of Stable Diffusion 3 (SD3)~\cite{sd3} has led us to reconsider the subject of part-level generation. 
The starting point for our paper is the observation that SD3 can capture part-level correspondences significantly better than prior models (as shown quantitatively in Table~\ref{tab:2d_diffuison} and qualitatively in Figure~\ref{fig:diffusion_comparison}). 
This new capability allows for the generation of complex part-aware entities through part-specific text descriptions. However, such fine-grained understanding capabilities are not yet available for 3D generation.

As SDS is potentially capable of lifting any entities from 2D to 3D, a straightforward approach might combine SD3 with SDS. However, as shown in Figure~\ref{fig:sd3_cross_attn}, we observe that SD3 occasionally struggles to generate animals according to prompts that specify particular animal body parts, even though it understands where those parts should be in the cross attention maps. This issue arises because part-correspondence understanding is only reliable at certain timesteps and transformer layers, making the SDS process less robust to prompts that focus on specific parts. 
Additionally, naively using SD3 within SDS is leading to multiple issues. SD3’s use of the rectified flow match Euler discrete scheduler results in deformed outputs with standard timestep sampling used for other diffusion models, as seen in Figure~\ref{fig:intro_demo} and~\ref{fig:baseline_comparison}. 
Other issues associated with SDS such as multi-face Janus problem and content drift~\cite{mvdream} are also present.
Furthermore, generating 3D assets with SD3 takes 7 hours, which can be prohibitive in many applications.

\begin{figure*}[t!]
    \centering
    \includegraphics[width=0.92\textwidth]{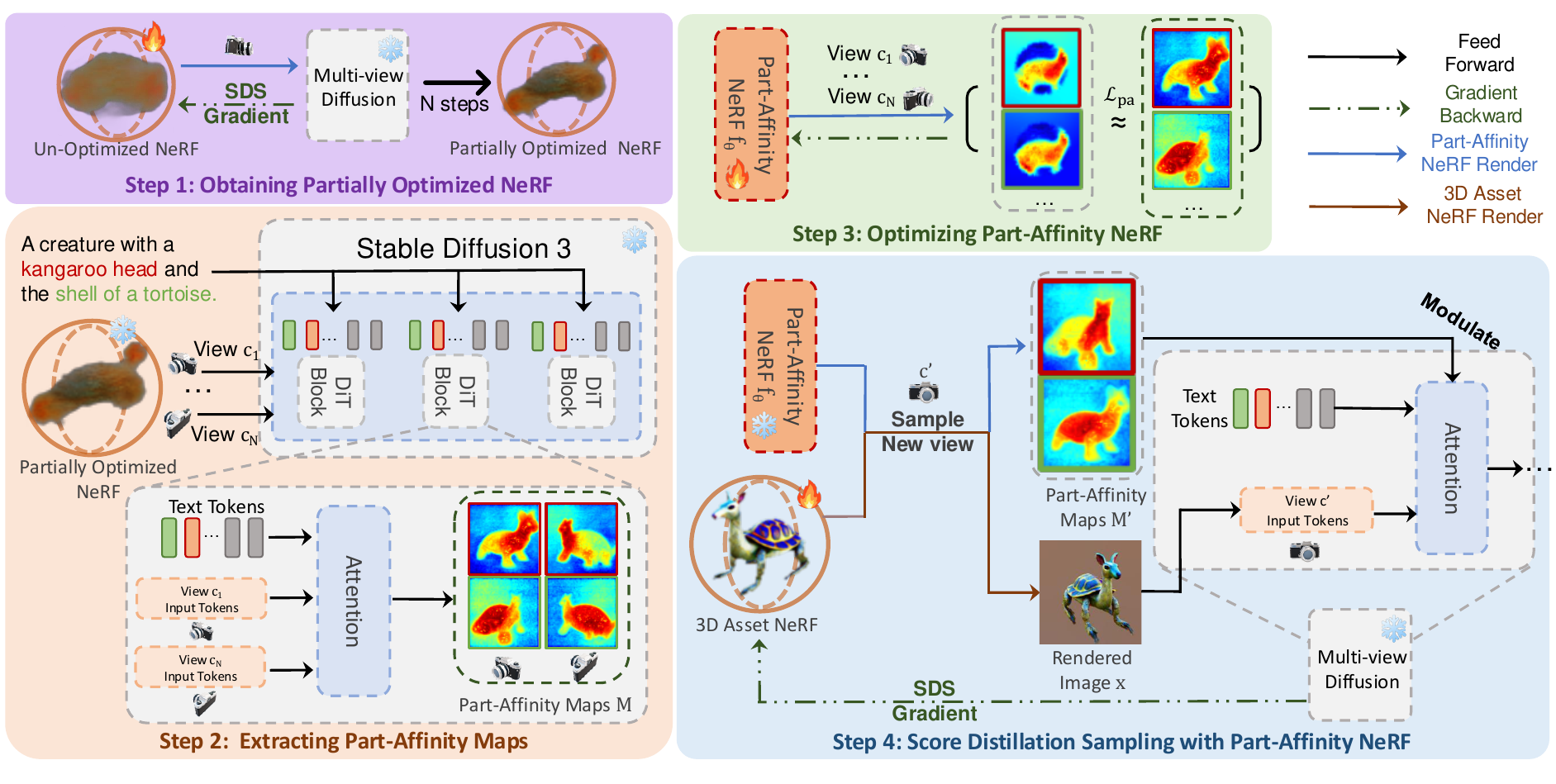}
    \caption{\textbf{\mname~pipeline.} \mname~is composed of 4 steps: (1)Partially optimize a NeRF under standard SDS. (2) Multiple rendered views of the partially optimized NeRF are input into SD3 along with the text prompt to construct Part-Affinity maps based on cross-attention of part tokens in SD3. (3) A Part-Affinity NeRF is trained using these extracted maps. (4) Both the trainable 3D asset NeRF and frozen Part-Affinity NeRF are rendered from the same camera pose. The rendered Part-Affinity map then modulates cross and self-attention in MVDream for SDS, ultimately generating a part-aware 3D animal.
    The symbol \snowflake~denotes a frozen model, while \fire~indicates a model that is trainable.
}
    \label{fig:method}
\end{figure*}
To overcome the aforementioned challenges, we introduce \mname, a novel part-aware knowledge transfer module designed to efficiently distill part-level understanding from powerful single-view diffusion models, such as SD3, and inject it into 3D generation with SDS.

\mname~first performs SDS over several steps to partially optimize the NeRF, producing a coarse yet view-consistent layout of the animal’s shape.
We then render the NeRF from a limited number of camera viewpoints.
We use these renderings as the denoising condition for SD3 and perform several denoising steps during which we extract cross-attention maps of part-specific tokens from part-aware layers.
We average these cross-attention maps for each of the camera viewpoints and obtain part affinity maps.

Next, we train a 3D Part-Affinity representation based on NeRF from the extracted part affinity maps, which allows us to interpolate part affinity maps from any camera viewpoint almost instantaneously.
Subsequently, during SDS, we render both the 3D asset and the learned Part-Affinity NeRF of \mname~from the same camera perspective and modulate the cross and self-attention mechanisms of the guidance model using the rendered part affinity map. 
This modulation causes regions with high part affinity to have higher responses to part-specific prompts. 
Consequently, our approach (\mname) not only promotes more reliable part-aware SDS but also significantly reduces the computational cost from 7 hours to 78 minutes and cuts GPU memory usage by 24GB, compared to the combination of SD3 with SDS as demonstrated in Figure~\ref{fig:running_speed}.
Quantitative and qualitative evaluations of \mname~ demonstrate its exceptional ability to generate part-aware, imaginative 3D animals.

In summary, our \textbf{main contributions} are as follows:
\begin{enumerate}
    \item We are the first to consider the problem of part-level text to 3D generation in an open-world setting. 
    \item We propose \mname, a novel knowledge transfer mechanism that efficiently transfers part-level knowledge of a 2D diffusion model into the 3D generation process. 
    \item We significantly improve the quality and decrease the computational cost of creating part-aware 3D animal assets by integrating \mname~within the SDS optimization process. 
    \item We demonstrate through quantitative evaluations and a human study that our method consistently outperforms baseline methods.
\end{enumerate}

\section{Related Work}
\paragraph{Lifting 2D Diffusion Models for 3D Animal Generation.}
Due to the limited generalizability of current 3D generative models, efforts have been made to adapt 2D diffusion priors or single image/video~\cite{learning3dfauna, wu2023magicpony} for 3D assets such as animals. The distilling diffusion prior approach primarily employs score distillation sampling (SDS)~\cite{dreamfusion}, where 2D diffusion priors serve as score functions that guide the optimization of 3D structures. Similarly, SJC~\cite{sjc} utilized publicly accessible diffusion models for their method. Subsequent studies have aimed at refining 3D representations, enhancing loss design, or implementing multi-stage optimizations~\cite{tang2023dreamgaussian, GaussianDreamer, magic3d, gsgen, wang2023prolificdreamer, huang2023dreamwaltz, yan2024dreamdissector, cheng2023progressive3d}. Some methods~\cite{melaskyriazi2023realfusion, jakab2023farm3d, Magic123, sun2023dreamcraft3d} leverage diffusion guidance to optimize 3D models based on a single image. Another set of methods uses diffusion guidance to learn the layout for compositional generation~\cite{chen2024comboverse, epstein2024disentangled3dscenegeneration, po2024compositional}. Notably, MVDream~\cite{mvdream} proposed a multi-view consistent diffusion model for SDS guidance, significantly improving issues with multi-face Janus and content drift.

\paragraph{Layout Guided 3D Generation.}
Earlier research~\cite{niemeyer2021giraffe} has explored the application of compositional neural radiance fields within an adversarial learning framework to achieve 3D-aware image generation. A pioneering study~\cite{funkhouser2004modeling} utilized a mesh database to find and combine parts to create new objects. Subsequent research incorporated probabilistic models for part suggestion~\cite{kalogerakis2012probabilistic}, semantic attributes~\cite{chaudhuri2013attribit}, and fabrication~\cite{schulz2014design}. Some studies~\cite{tertikas2023generating} employed neural radiance fields to represent various 3D elements and render these into a unified 3D model. Recent advancements ~\cite{po2024compositional}, guided by pre-trained diffusion models, have enabled the generation of compositional 3D scenes using user-provided 3D bounding boxes and text prompts. Concurrently, other works have used large language models (LLMs) to generate 3D layouts from text prompts as an alternative to human annotations~\cite{gao2024graphdreamer, vilesov2023cg3d, wang2023luciddreaming}, or combined layout learning during the optimization process~\cite{chen2024comboverse, epstein2024disentangled3dscenegeneration}. 
While these approaches can produce 3D scenes through composition, they all rely heavily on scene graphs or descriptions of object-to-object relationships for object-to-scene generation.
This dependency makes it impossible when it comes to composing part-to-object generational tasks.
Unlike object-to-scene generation, which requires an understanding of the relationships between distinct objects, part-to-object generation necessitates a more fine-grained comprehension of how individual parts combine to form a coherent whole.

\paragraph{Diffusion with Cross-Attention Control.} 
Since most current diffusion models are transformer-based and incorporate text information through cross-attention layers, providing spatial awareness naturally aligns with cross-attention control. Several studies~\cite{chen2023trainingfree, hertz2022prompt, densediffusion, Phung2023GroundedTS, epstein2023selfguidance} explore various methods to enhance cross-attention scores between regions and their corresponding descriptions in the prompts. In contrast, others~\cite{He2023LocalizedTG} propose applying a binary mask to eliminate attention between regions and non-matching region descriptions. To the best of our knowledge, cross-attention control is predominantly applied in 2D generation models and is seldom utilized in 3D SDS settings as in our paper.

\begin{figure*}[h!]
    \centering
    \includegraphics[width=0.83\textwidth]{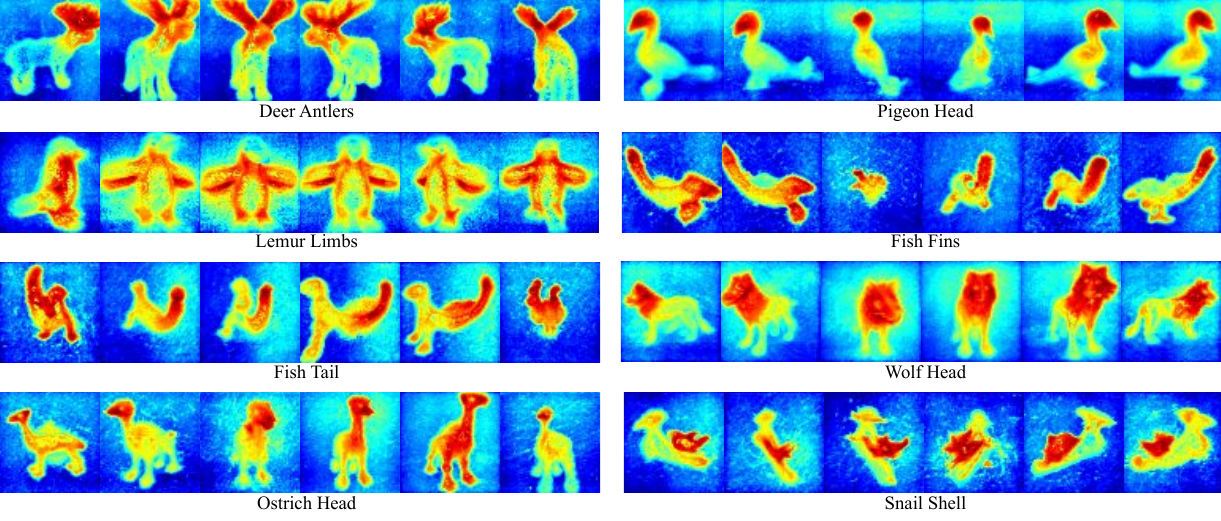}
    \caption{\textbf{Learned  part affinity map rendered from unseen camera poses.} Heatmaps displaying the learned 3D  part affinity representation rendered from unseen camera poses for different part-specific descriptions of distinct animals. Warmer colors indicate stronger affinities, highlighting our implicit 3D neural representation's capability to differentiate and localize specific anatomical features.}
    \label{fig:part_affinity_map}
\end{figure*}
\section{Method}

To efficiently transfer part-level knowledge of a 2D diffusion model into the 3D generation process, we introduce \mname~as a novel geometry-consistent mechanism designed for this purpose.
In Section~\ref{sec:method-prelim}, we revisit the classic Score Distillation Sampling (SDS) formulation and discuss the issues that arise when applying SDS directly with SD3. 
Subsequent sections detail our method: Section~\ref{sec:method-part_map_extract} describes the extraction of 2D part affinity maps from SD3, Section~\ref{sec:method-3Drep} details the construction of our Part-Affinity NeRF using the part affinity maps, and Section~\ref{sec:method-sds} presents the integration of the Part-Affinity NeRF within SDS to generate part-aware 3D assets.

\subsection{Preliminaries}\label{sec:method-prelim}
Before delving into our method, we briefly review the concepts commonly employed in the 3D lifting generation techniques and further discuss our motivation.

\paragraph{Score Distillation Sampling.}
As introduced by~\cite{dreamfusion}, Score Distillation Sampling (SDS) uses a pre-trained 2D diffusion model with fixed parameters $\phi$ to guide the generation of 3D models with the vast amount of 2D image prior knowledge. 
Let $\theta$ denote the 3D representation such as NeRF~\cite{nerf} or Gaussian Splatting~\cite{gs} and $g(\cdot)$ be the differentiable rendering function, which renders the 3D model to an image $x = g(\theta)$. 
During the guidance process with $y$ as the text condition, we first sample a random timestep $t \sim \{0,..., T\}$ and a random noise $\epsilon \sim \mathcal{N}(0, I)$.
We then add the noise to the rendered image and we get $x_t = \sqrt{\overline{\alpha_t}} x + \sqrt{1 - \overline{\alpha_t}} \epsilon$.
The SDS gradient then is defined as follows:
\begin{equation*}
\nabla_{\theta} \mathcal{L}_{SDS}(x = g(\theta)) = \mathbb{E}_{t,\epsilon} \left[ w(t) (\hat{\epsilon}(x_t; y, t, \phi) - \epsilon) \frac{\partial x}{\partial \theta} \right]
\end{equation*}
where $\overline{\alpha_t}$ and $w(t)$ are weighting functions that depend on the timestep t, and $\hat{\epsilon}$ is the predicted noise by the pre-trained diffusion model.

\paragraph{Why not Stable Diffusion 3 as the Guidance Directly?} \label{sec:why_not_sd3}
A straightforward way to leverage part-level knowledge from SD3 is by performing SDS with SD3’s guidance. However, we argue that this approach is ineffective for several reasons: (1) Even when SD3 manages to generate images with part-level specifications, it often fails because part-level understanding is only exhibited at specific transformer layers and timesteps, as shown in Figure~\ref{fig:sd3_cross_attn}. During the denoising forward pass through SD3, information about the animals and their parts can become mixed, leading to a loss of part-level control. This makes SDS unstable and unpredictable at the part level. (2) SD3 cannot provide view-consistent guidance, which leads to issues like multi-face Janus problem and content drift problems~\cite{mvdream}. (3) SD3 uses a rectified flow match Euler discrete scheduler, which differs from previous diffusion methods. We observe that timestep sampling strategies from earlier methods do not yield satisfactory results with SD3. Additionally, hyperparameters such as the guidance scale and 3D shape loss scales require extensive empirical tuning. (4) The forward process is computationally expensive, requiring 48GB of GPU memory and 7 hours of training time on an NVIDIA A40 to generate just one 3D asset.
In contrast, \mname~ offers more stable and robust part-level specification during SDS and requires only 78 minutes to complete. This is slightly longer than SD2.1 or MVDream, which takes 50 minutes on an A40 GPU (see Figure~\ref{fig:running_speed} for a comparison of running speeds).

\paragraph{Why not Stable Diffusion 3 + MVDream as the Guidance?}
This approach could potentially solve the multi-face Janus problem; however, the remaining issues with SD3 still persist.
Moreover, this combination requires 58GB of GPU memory and takes 8 hours to generate a single 3D asset through SDS, rendering it highly inefficient.

\subsection{Part Affinity Map Extraction} \label{sec:method-part_map_extract}
Before extracting the part affinity map from SD3, we first perform SDS for several steps to partially optimize the NeRF. We then render this partially optimized NeRF from various camera angles to obtain view-consistent, coarse animal-shape layouts. These view-consistent layouts serve as conditions for SD3, where the rendered animal shape is mixed with noise as input for denoising, which helps keep the extracted part affinity maps view-consistent as well.

We choose Stable Diffusion 3 (SD3) as our source of 2D part-level knowledge for two key reasons: (1) Unlike perception-driven part-level segmentation frameworks~\cite{ovparts, catseg, zsseg, car} that rely on well-defined input images,
we found that SD3 can operate effectively on noisy images generated from partially optimized NeRFs.

(2) Among many open-source models we examined, SD3 is the only one that demonstrates part-level understanding in its cross-attention maps.

To obtain the part affinity map, we conduct a denoising process for the timesteps between $t_{s}$ and $t_{e}$.
Specifically, for a rendered image $x$ from a partially optimized NeRF under camera pose $c$, we introduce noise to $x$ using a weighting factor $\alpha_{t_{s}}$ to produce $x_{t_{s}}$.
Subsequently, we perform denoising in the latent space for each timestep up to $t_{e}$.
At each timestep $t$, and for each transformer layer $l$ in SD3, we compute an attention map $A_{t, l, c} \in \mathbb{R}^{HW \times n}$, where $n$ denotes the number of tokens in the text prompt $y \in \mathbb{R}^n$, and $HW$ represents the spatial resolution of the feature maps.
From this attention map, we extract a spatial correspondence map $M_{t, l, i, c} \in \mathbb{R}^{HW}$ for each token $y_i$, which corresponds to a slice of $A_{t, l, c}$ associated with the token $y_i$.
Let $I_p$ represent the set of token indices for the part-level description in $y$, for example, "kangaroo head" in "a creature with a kangaroo head and a tortoise shell" corresponds to a set of token indices $I_p = \{4, 5\}$.
We can then derive the part affinity map $M_{p,c} \in \mathbb{R}^{HW}$ for this camera pose using the following equation:
\begin{equation*}
    M_{p, c} = \frac{1}{(t_s - t_e) \cdot L \cdot |I_p|} \sum_{t=t_s}^{t_e} \sum_{l=0}^{L} \sum_{i\in I_p} M_{t, l, i, c}
\end{equation*}
where $L$ is the number of transformer layers.

\subsection{{Part-Affinity NeRF}} \label{sec:method-3Drep}
\begin{figure}[h!]
    \centering
    \includegraphics[width=0.48\textwidth]{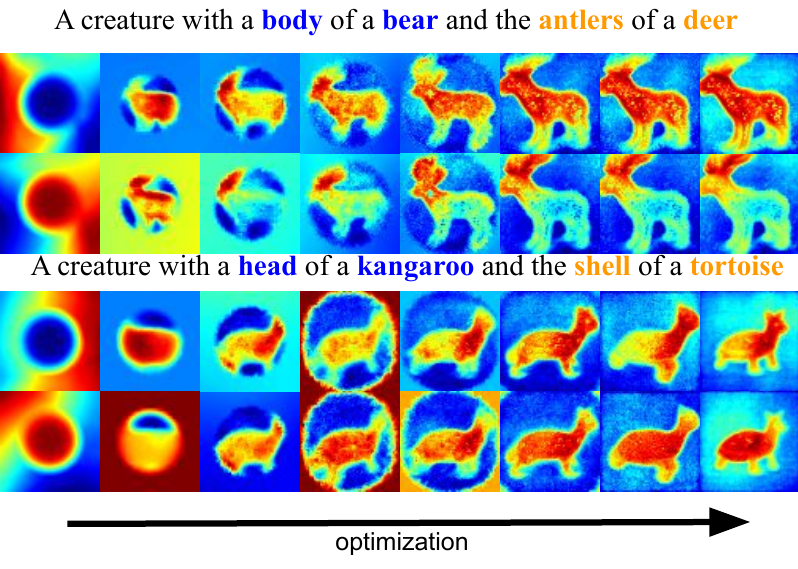}
    \caption{\textbf{Part-Affnity NeRF learning progress visualization.}
    We show the evolution of renderings of the Part-Affinity NeRF for each corresponding part (e.g., body of a bear in the first row) throughout the learning, demonstrating that the Part-affinity NeRF quickly converges.}
    \label{fig:part_affinity_nerf}
\end{figure}
The part affinity map $M_{p, c}$ applies only to a specific camera pose $c$. 
This poses a problem, as the camera poses are sampled from a continuous distribution during SDS, implying an infinite number of potential camera poses.

A naive approach would be to generate the part affinity map each time an image is rendered from the 3D asset.
However, this approach significantly increases computational demands.
The forward pass in SD3, combined with the need to obtain cross-attention maps for every layer and timestep, requires substantially more computation.
As a result, generating a single 3D asset can take up to 58 hours, which is even longer than using SD3 solely as a guidance mechanism for SDS.

Therefore, as part of \mname, we introduce the Part affinity NeRF that learns from part affinity maps of discrete camera poses and is capable of interpolating these maps for continuous camera poses.

Mathematically, let $\{\{M_{p,c}\mid p \in \mathcal{P}\} \mid c \in \mathcal{C}\}$ denote all part affinity maps obtained for all parts $\mathcal{P}$ under a set of camera poses $\mathcal{C}$. The optimization goal is to fit the implicit representation $  f_\theta (M\mid p, c)$ parameterized by a neural network with parameters \( \theta \) :
\begin{equation*}
    \min_{\theta} \mathcal{L}_{pa} =  \min_{\theta}\sum_{c \in \mathcal{C}}\sum_{p\in \mathcal{P}}||f_\theta (M\mid p, c) - M_{p, c}||^2
\end{equation*}

In our design, $f_\theta$ is an MLP-based neural radiance field. 
The continuous nature of MLPs imposes a form of smoothness and continuity in the learned representation, enabling \mname~ to produce part affinity maps from any camera pose. 
In conclusion, \mname~ offers several key benefits:
(1) it provides a 3D consistent representation of part affinity maps learned from discrete camera views, allowing for the interpolation of part affinity maps under any camera pose, which makes it efficient to use with the SDS process;  
(2) it significantly reduces the computation cost of using SD3 at every single step to the more economical learning and rendering cost of \mname, reducing the training time from 58 hours to 78 minutes per 3D asset.

\begin{algorithm}

\caption{Layer-wise Attention Modulation with \mname}
\begin{algorithmic}[1]
\Require Cross attention and self attention of the 2D diffusion guidance model $\mathcal{S}_{cross}, \mathcal{S}_{self} $, rendered part affinity map $\{f_\theta (M\mid p, c')| p\in \mathcal{P}\} $, part token indices $\{I_p \mid p\in \mathcal{P}\}$, enhancement factor $\alpha_{cross}$, $\alpha_{self}$
\For{$p$ in $\mathcal{P}$}  \Comment{Cross Attention}
    \State $M'_{p,c'} \gets f_\theta (M\mid p, c')$
    \State $\mathcal{S}_{cross}[:, I_p] \gets \mathcal{S}_{cross}[:, I_p] + \alpha_{cross} \log M'_{p, c'}$
    \State $A_{cross} \gets \text{Softmax}(\mathcal{S}_{cross})$ 

\EndFor

\For{$p$ in $\mathcal{P}$}  \Comment{Self Attention}
        \State $M'_{p,c} \gets f_\theta (M\mid p, c')$
        \State $K_{p, c'} \gets (M'_{p, c'})^TM'_{p, c'}$ \Comment{Symmetry}
        \State $\mathcal{S}_{self} \gets \mathcal{S}_{self} + \alpha_{self}\log K_{p, c'}$
        \State $A_{self} \gets \text{Softmax}(\mathcal{S}_{self})$
\EndFor

\end{algorithmic}
\label{algo-attn}
\end{algorithm}

\subsection{SDS with Attention Modulation} \label{sec:method-sds}
After rendering an image \( x(c') \) from the 3D asset under a specific camera pose \( c' \) during SDS, we also generate the rendered affinity maps for each of the parts \( \{f_\theta (M\mid p, c')| p \in \mathcal{P}\} \) using the optimized Part-Affinity NeRF under the same camera pose.
These rendered part affinity maps are utilized to modulate the cross-attention and self-attention matrices in the 2D diffusion guidance model for SDS.
Specifically, we modulate the cross-attention score maps \( \mathcal{S}_{cross} \in \mathbb{R}^{hw \times n} \) and the self-attention score maps \( \mathcal{S}_{self} \in \mathbb{R}^{hw \times hw} \) (before the softmax operation), where $hw$ represents the feature spatial resolution and $n$ denotes the number of tokens, in the 2D diffusion guidance models at each denoising step \( \hat{\epsilon}(x_t; y, t, \phi) \).
The detailed procedure is outlined in Algorithm~\ref{algo-attn}.
The cross-attention modulation ensures that regions corresponding to a specific part are guided by their corresponding part-specific token.
The self-attention modulation increases influence within intra-part regions and reduces influence among inter-part regions.

\section{Experiments}
\begin{table}[t!]
\centering 
\tablestyle{1.6pt}{1.0}
\scalebox{1}{
\begin{tabular}{lccc}
\toprule
2D Diffusion Model& 2 Parts & 3 Parts \\
\midrule
MVDream~\cite{mvdream}  & 0.242 & 0.108 \\

Stable Diffusion 2.1~\cite{sd21}  & 0.187 & 0.022 \\


Stable Diffusion XL & 0.297 & 0.032 \\


DeepFloyd~\cite{deepfloyd}  & 0.429 & 0.097 \\

 \rowcolor{Gray!16} Stable Diffusion 3~\cite{sd3} & \textbf{0.826} & \textbf{0.537} \\



\bottomrule
\end{tabular}}
\caption{\textbf{2D part-aware generation success rate. }A user study involving five participants found that SD3 has a significantly higher success rate than other popular diffusion models in generating part-aware images based on part-level prompts (describing 2 or 3 animal parts in a single prompt). This suggests that SD3 has a superior ability to understand and generate images at the part level.}
\label{tab:2d_diffuison}
\end{table}

\begin{table}[]
    \centering
    \tablestyle{1.6pt}{1.0}
    \scalebox{0.85}{
    \begin{tabular}{lccc}
        \toprule
        \multirow{2}{*}{Method} & \multicolumn{3}{c}{CLIP Score$\uparrow$} \\
        \cmidrule{2-4}
         &B/32 & B/16 & L/14 \\
        \midrule
        DreamFusion(SD2.1)~\cite{dreamfusion} &  0.271$\pm 3.0e^{-4}$&  0.274 $\pm 2.2e^{-4}$ &   0.226$\pm 3.6e^{-4}$\\
        DreamFusion(SD3)~\cite{dreamfusion}&  0.271$\pm 5.8e^{-4}$& 0.275$\pm 5.5e^{-4}$& 0.229$\pm 7.7e^{-4}$\\
        MVDream~\cite{mvdream} &  0.275$\pm 7.8e^{-4}$&  0.282$\pm 4.1e^{-4}$&    0.230$\pm 7.9e^{-4}$\\
        GeoDream~\cite{GeoDream} &  0.244$\pm 1.2e^{-4}$&  0.252 $\pm 1.5e^{-4}$   &    0.202$\pm 1.8e^{-4}$\\
        OpenLRM~\cite{hong2023lrm} &0.265 $\pm 2.1e^{-4}$ & 0.285 $\pm 2.4e^{-4}$ & 0.223 $\pm 6.2e^{-4}$     \\
        VFusion3D~\cite{han2024vfusion3d} &  0.268 $\pm 1.9e^{-4}$&   0.281 $\pm 2.1e^{-4}$  &  0.225 $\pm 6.2e^{-4}$  \\
         \rowcolor{Gray!16} \mname~(Ours)  &  \textbf{0.285$\pm 2.6e^{-4}$}& \textbf{0.289$\pm 3.5e^{-4}$}& \textbf{0.245$\pm 4.6e^{-4}$}\\
        \bottomrule
    \end{tabular}}
   \caption{\textbf{Performance comparison of different methods.}} Our method shows the best CLIP scores among all.
    \label{tab:performance_comparison}
\end{table}

\begin{table}[]
    \centering
    \tablestyle{1.6pt}{1.0}
    \scalebox{1}{
    \begin{tabular}{lccc}
        \toprule
        \multirow{2}{*}{View Number} & \multicolumn{3}{c}{CLIP Score$\uparrow$} \\
        \cmidrule{2-4}
         &B/32 & B/16 & L/14 \\
        \midrule
        8& 0.274$\pm 5.0e^{-4}$&  0.281$\pm 4.6e^{-4}$&  0.231$\pm 5.9e^{-4}$ \\
        16& 0.277$\pm 4.8e^{-4}$& 0.281$\pm 1.9e^{-4}$& 0.232$\pm 5.8e^{-4}$\\
        32 & 0.277$\pm 4.5e^{-4}$& 0.283$\pm 4.3e^{-4}$& 0.235$\pm 5.7e^{-4}$  \\
        64& 0.284$\pm 4.3e^{-4}$& 0.287$\pm 4.0e^{-4}$& 0.240$\pm 5.5e^{-4}$  \\
       76& 0.285$\pm 2.6e^{-4}$& 0.289$\pm 3.5e^{-4}$& 0.245$\pm 4.6e^{-4}$ \\

        \bottomrule
    \end{tabular}}
    \caption{\textbf{Ablation study on the number of views for extracted part-affinity maps.} Increasing the number of views results in a stronger part-affinity NeRF.}
    \label{tab:ab_view}
\end{table}

\begin{figure*}[h!]
    \centering
    \includegraphics[width=0.96\textwidth]{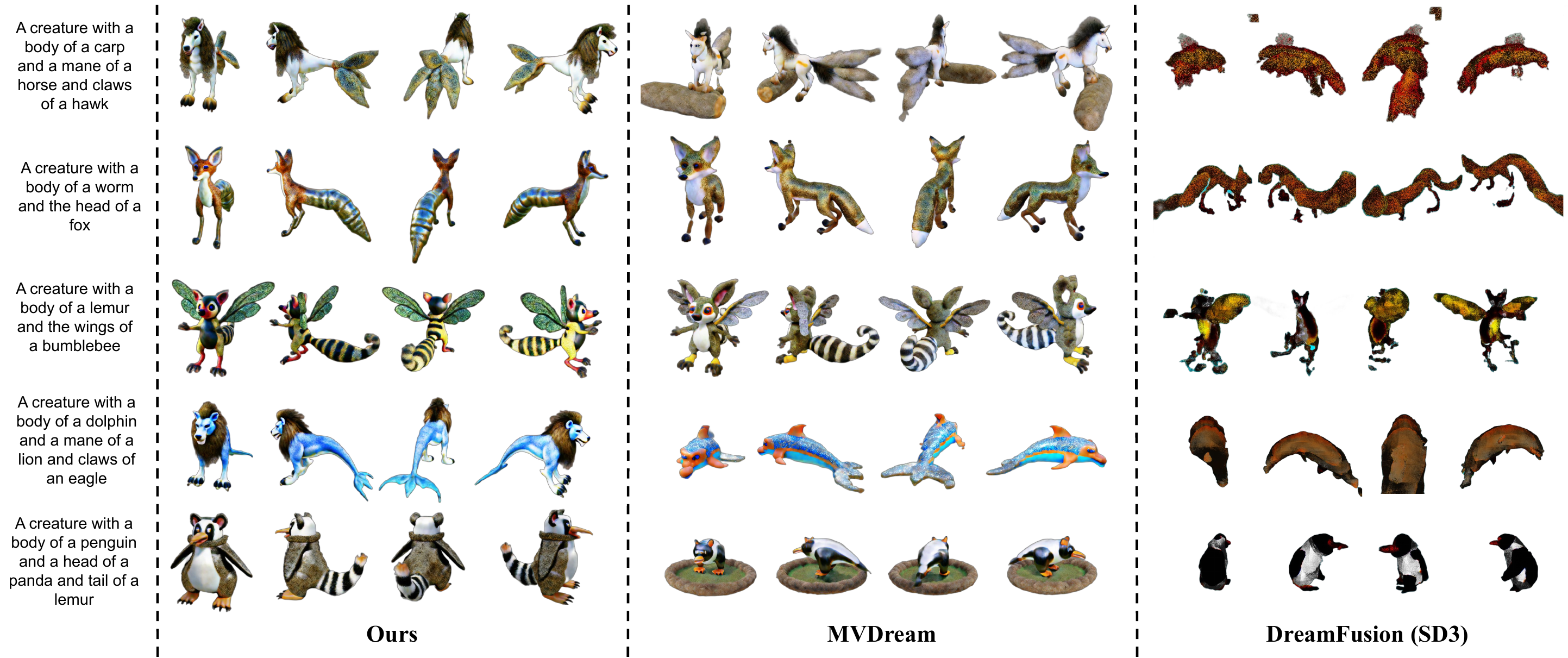}
    \caption{\textbf{Qualitative results of generated fantastic animals from different methods.} \mname is capable of generating 3D assets with better part correspondence to part-specific prompts compared to MVDream or SD3. The fantastic animals created by MVDream and SD3 often either omit certain body parts or blend different animal elements globally, which is not the desired outcome.
}
    \label{fig:baseline_comparison}
\end{figure*}
\subsection{Implementation Details}
 
We consider the cross-attention map between $t_s=450$ and $t_e=100$ at the eleventh layer ($l=11$) in SD3, where we found the most significant part-level understanding.

Our Part-Affinity NeRF is a small MLP with one hidden layer comprising 16 neurons.
The output dimension is equivalent to the number of parts described in the prompt, with each dimension representing a different part of the fantastical animal.
The part affinity map is rendered similarly to NeRF~\cite{nerf}, using 128 samples per ray and a rendered resolution of 64. 
We optimize Part-Affinity NeRF for 2000 steps.

We choose MVDream~\cite{mvdream} as our diffusion guidance model due to its multiview consistency and computational efficiency for SDS. 
We set $\alpha_{cross}=0.8$ and $\alpha_{self}=0.9$ to modulate the influence of cross and self-attention maps within MVDream. 
For the SDS optimization, we adhere to the default hyper-parameters specified in MVDream for other settings. The training procedure involves several steps to create a detailed 3D asset of an animal. 
Initially, the asset is trained (guided by MVDream) for 1000 steps to obtain the partially optimized NeRF. Following this, the Part-Affinity NeRF is optimized over an additional 2000 steps.
Finally, the training continues for another 4000 steps, using part-level guidance from the rendered part affinity maps, to achieve the final part-aware 3D model.

\subsection{Evaluation Benchmarks}

We use GPT-4o-mini to randomly generate 30 prompts under a template "a creature with a [animal 1][part 1], [animal 2][part 2], and [animal 3][part 3]". We use CLIP text similarity and ranking-based user study to evaluate how well the generated 3D assets match their descriptions. More visualizations and non-animal results are presented in Appendix~\ref{as:non-animal}.

\subsection{Main Results}
\paragraph{CLIP Similarity Experiment.} We compare \mname~with other distillation-based methods~\cite{mvdream, dreamfusion, GeoDream}, and popular feedforward 3D generation methods~\cite{hong2023lrm, han2024vfusion3d}, results are shown in Table~\ref{tab:performance_comparison}. We observe that \mname~consistently has higher similarity scores across CLIP types, indicating \mname's better part-correspondence in generated 3D assets and the part-specific prompts.

\paragraph{User Study Experiment.} 
\begin{figure}[h!]
    \centering
    \includegraphics[width=0.44\textwidth]{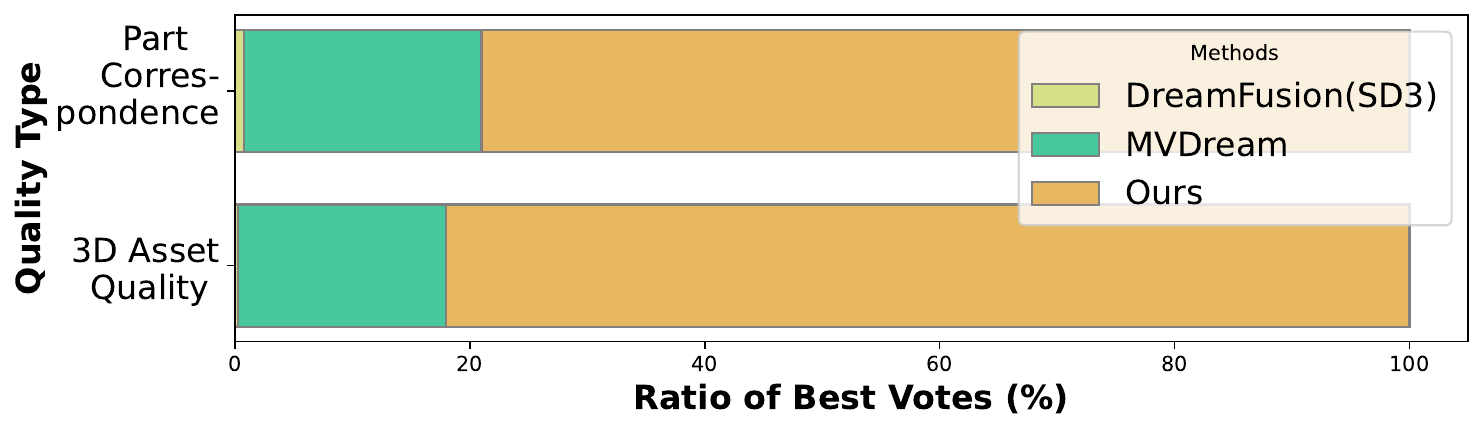}
    \caption{\textbf{User study results.} Participants were shown multi-view images generated from the same prompts and asked to select the best result. Our method receives significantly more votes for both part correspondence and overall 3D asset quality. More detail in the Appendix~\ref{as:human}}
    \label{fig:human_study}
\end{figure}
To assess the part correspondence quality and overall quality of the generated content, we asked users to compare our results with those from DreamFusion (SD3) and MVDream. We showed 24 users multi-view images generated from 20 random prompts and asked them to choose the best one. The results, shown in Figure~\ref{fig:human_study}, clearly demonstrate that \mname~outperforms existing methods. This suggests that our method effectively understands part compositions and generates high-quality 3D results.

\paragraph{Qualitative Results.}
As illustrated in Figure~\ref{fig:baseline_comparison}, \mname~demonstrates the ability to generate part-aware animals, with each relevant part closely adhering to its description.
In contrast, MVDream produces assets with globally mixed animal features.
Similarly, Dreamfusion (SD3) struggles to generate part-aware results, and both its 3D shape and texture quality are lacking, as discussed in Section~\ref{sec:why_not_sd3}.

\paragraph{Ablation Study.}
We examined the impact of the number of views on the extracted Part-Affinity maps, as shown in Table~\ref{tab:ab_view}. Increasing the number of views enhances the context of part affinity, enabling the optimized Part-Affinity NeRF to better interpolate under unseen camera poses. This results in more accurate part-knowledge integration during the SDS process, which improves the CLIP similarity score. Moreover, we observed that performance gains diminish beyond 64 views, suggesting that our method can effectively capture complete Part-Affinity information in 3D with a relatively small number of views. We provide additional analysis of our design choices in Appendix~\ref{as:ablation}.

\section{Conclusion}
This work incorporates Part-Affinity knowledge to address the challenges associated with a limited part-level understanding of SDS-based 3D asset generation methods.
Our proposed \mname~exhibits high precision in generating 3D assets with detailed part components, outperforming existing techniques in terms of both quality and efficiency.
This contributes to the advancement of the field of creative and complex 3D content creation, paving the way for the development of more detailed and imaginative digital worlds.

\paragraph{\textbf{Acknowledgments.}}
The authors would like to thank Paul Engstler for his insightful feedback on the manuscript.

{
    \small
    \bibliographystyle{ieeenat_fullname}
    \bibliography{main}
}
\clearpage
\setcounter{page}{1}
\begin{figure*}[t!]
    \centering
    \includegraphics[width=0.6\textwidth]{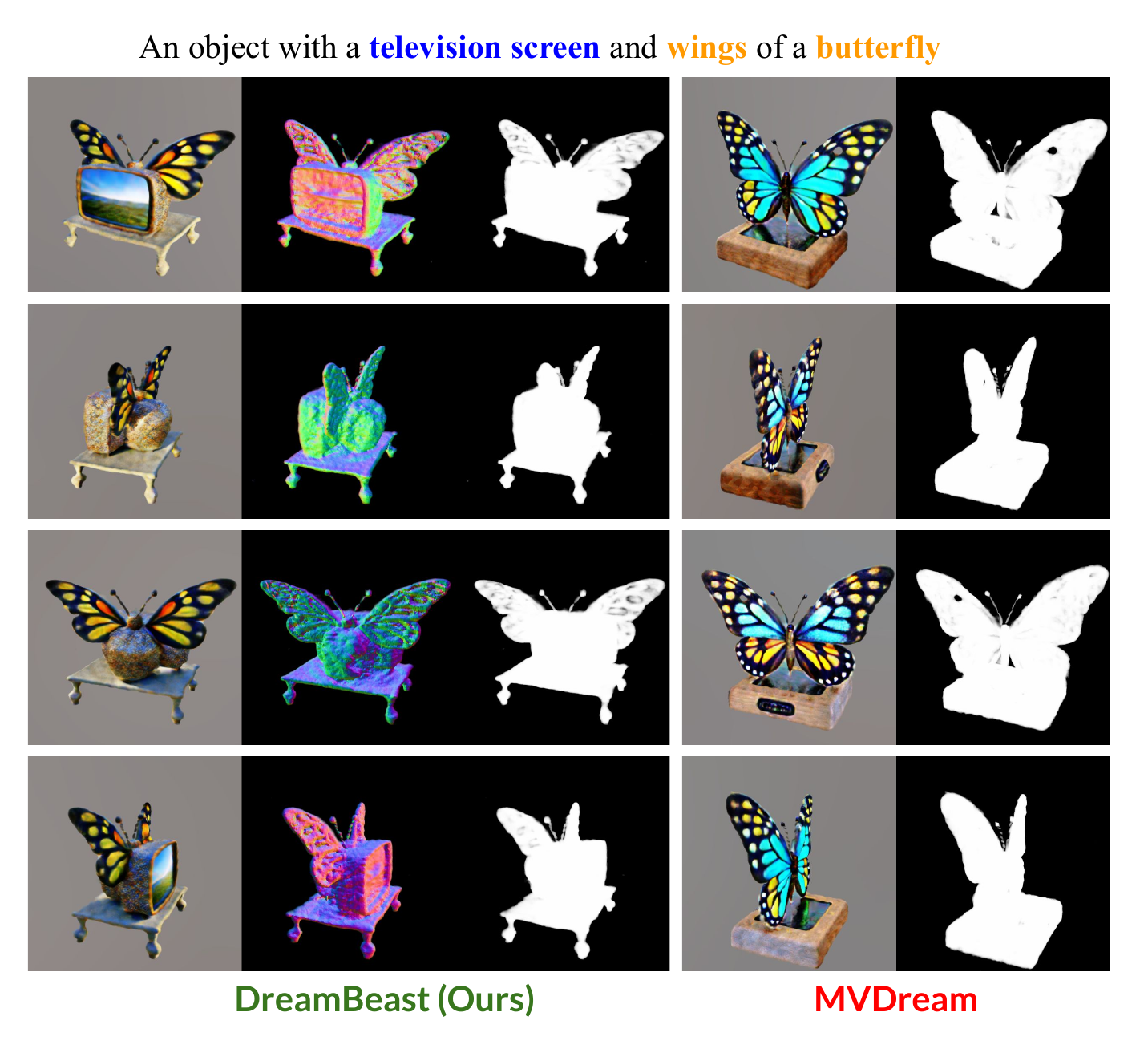}
    \caption{\textbf{Non-animal result generated by \mname} 
}
    \label{fig:non_animal_demo_0}
\end{figure*}
\begin{figure*}[t!]
    \centering
    \includegraphics[width=0.6\textwidth]{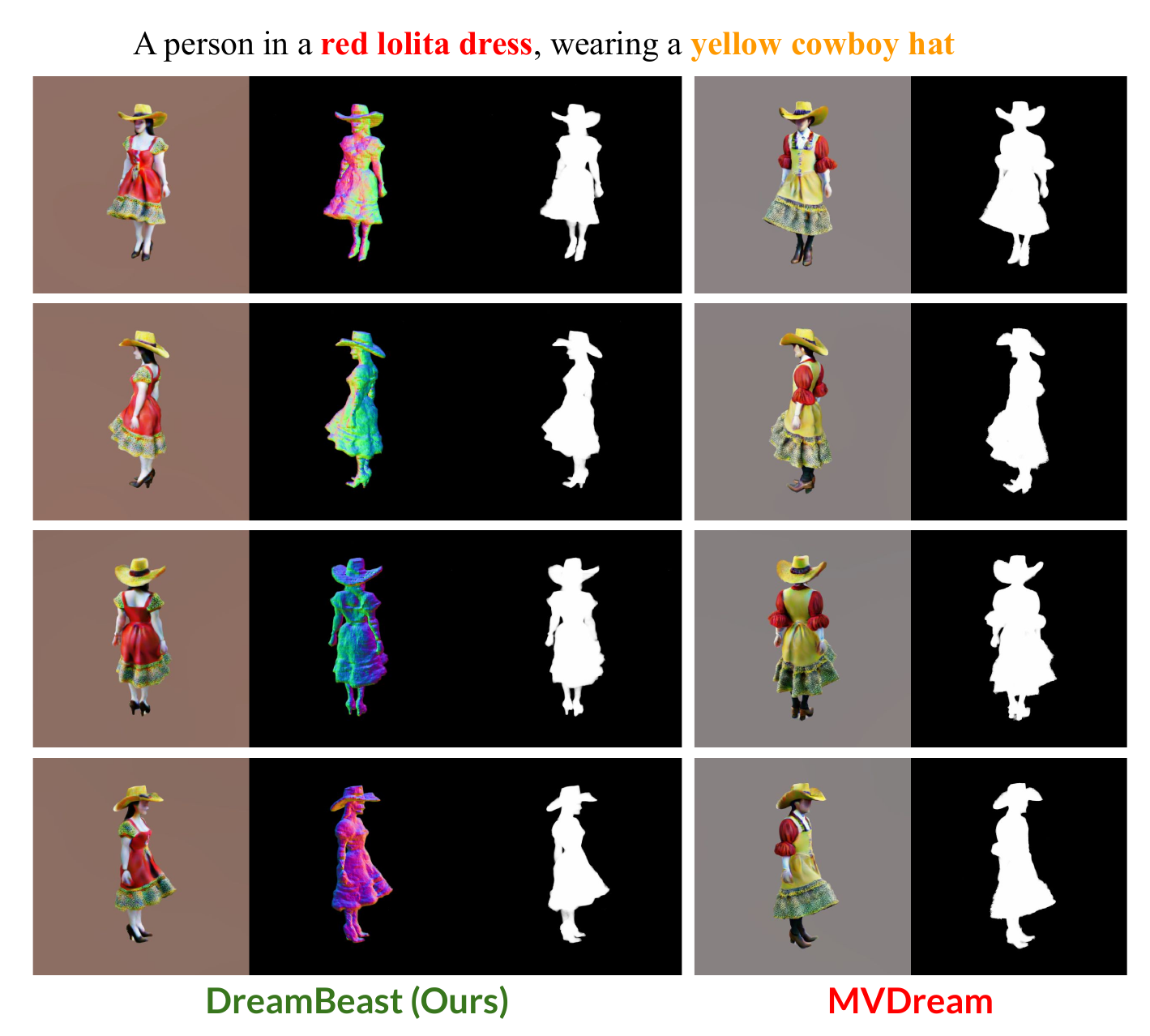}
    \caption{\textbf{Non-animal result generated by \mname} 
}
    \label{fig:non_animal_demo_1}
\end{figure*}
\begin{figure*}[t!]
    \centering
    \includegraphics[width=0.6\textwidth]{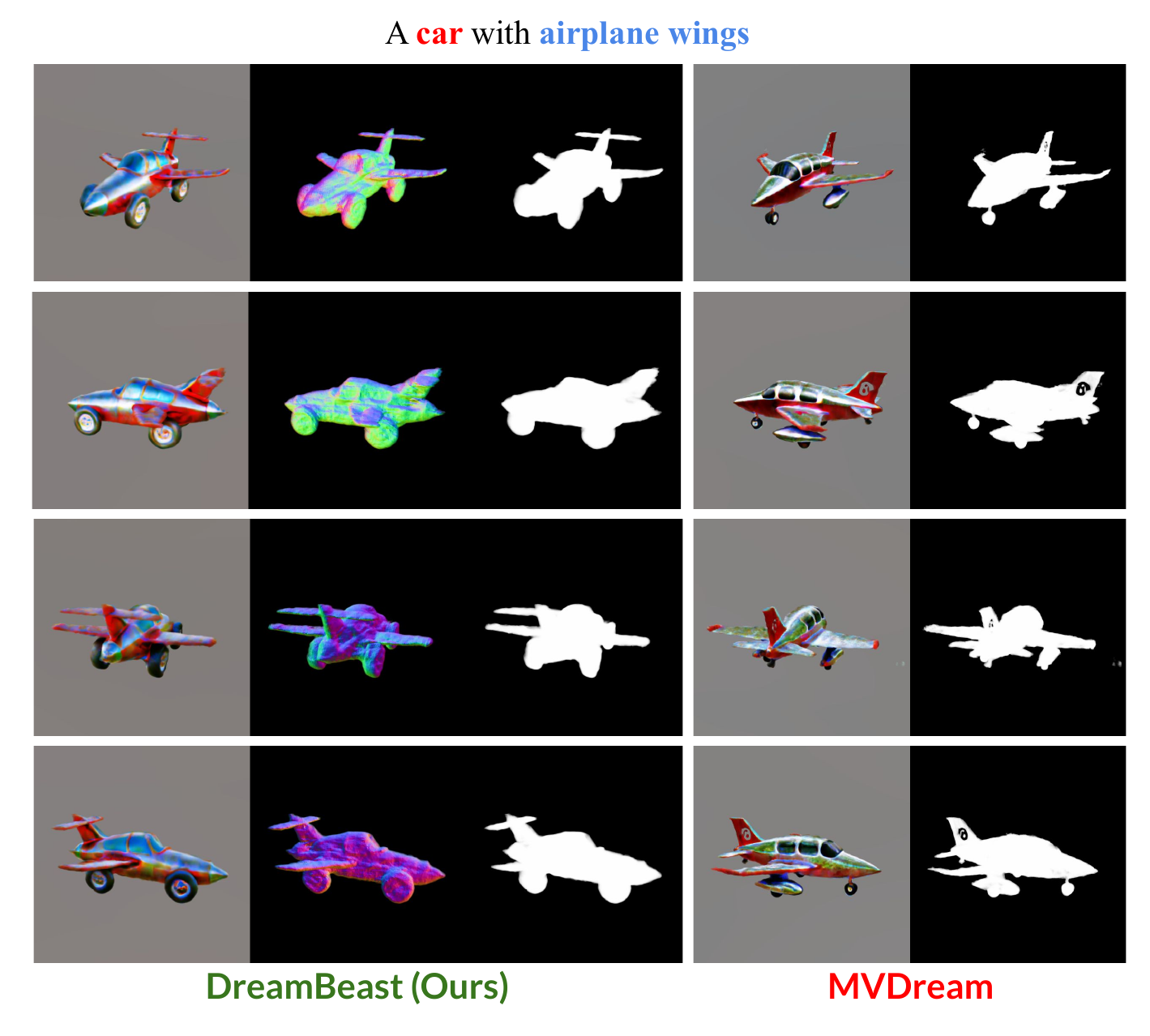}
    \caption{\textbf{Non-animal result generated by \mname} 
}
    \label{fig:non_animal_demo_2}
\end{figure*}
\begin{figure*}[t!]
    \centering
    \includegraphics[width=0.6\textwidth]{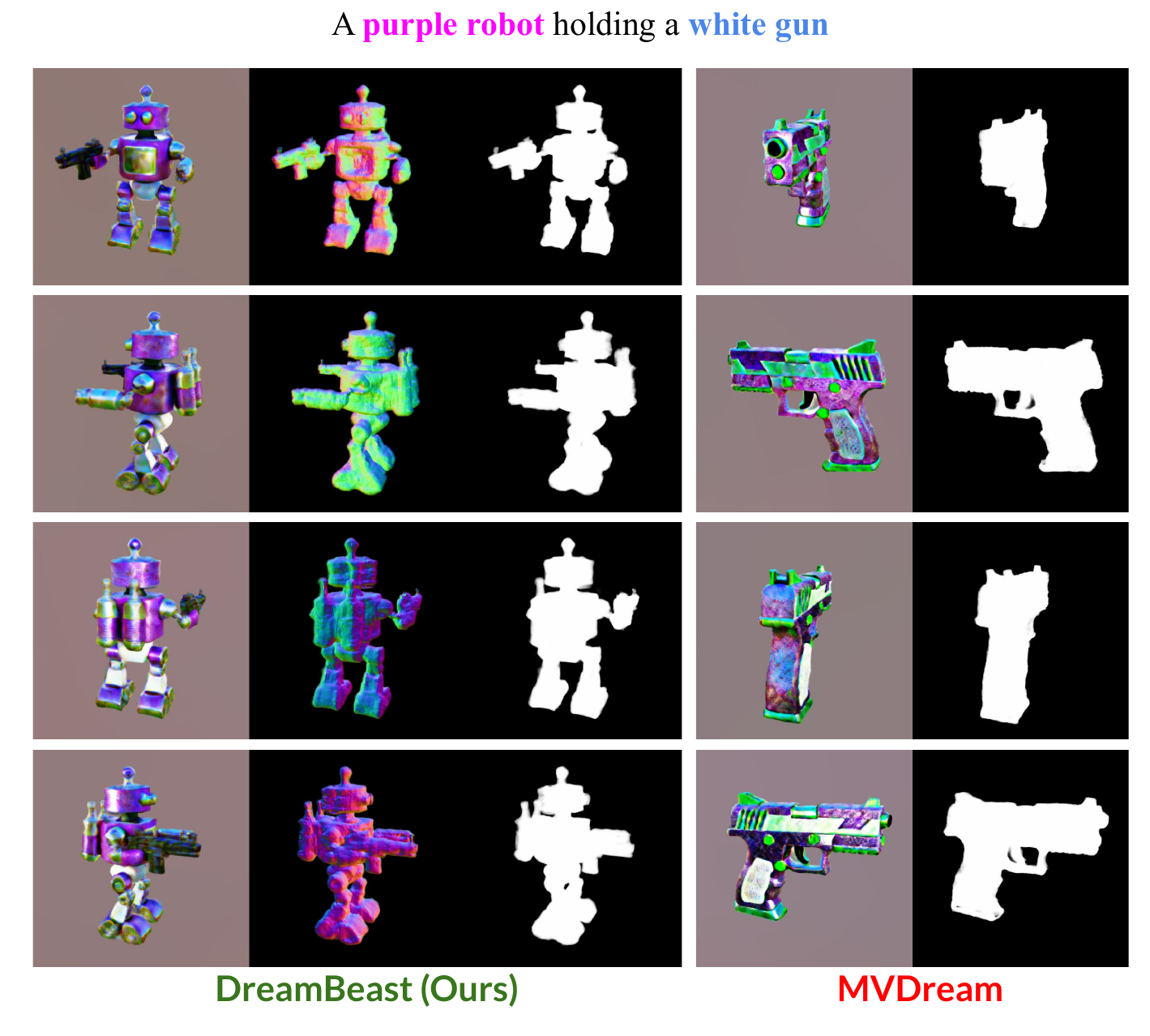}
    \caption{\textbf{Non-animal result generated by \mname} 
}
    \label{fig:non_animal_demo_3}
\end{figure*}
\appendix

\section*{Appendix}

This Appendix contains additional ablations (\cref{as:ablation}), more detailed quantitative results (\cref{as:human}), non-animal part-aware asset generated by \mname~compared to MVDream~\cite{mvdream} (\cref{as:non-animal}), running cost breakdown (\cref{as:speed}), more qualitative results of fantastic animals (\cref{as:qualitative}), failure case analysis (\ref{as:failure}), and more implementation details (\cref{as:implementation}).

\section{Additional Ablation Study}
\label{as:ablation}
We also conducted an ablation study on the cross-attention modulation factor ($\alpha_{cross}$) and self-attention modulation factor ($\alpha_{self}$), as detailed in Table~\ref{tab:ab_cross_factor}.
The performance of \mname~remains stable across a wide range of values for the modulation factors.

\section{Detailed Human Study Breakdown}
\label{as:human}
\begin{figure*}[h!]
    \centering
    \includegraphics[width=0.84\textwidth]{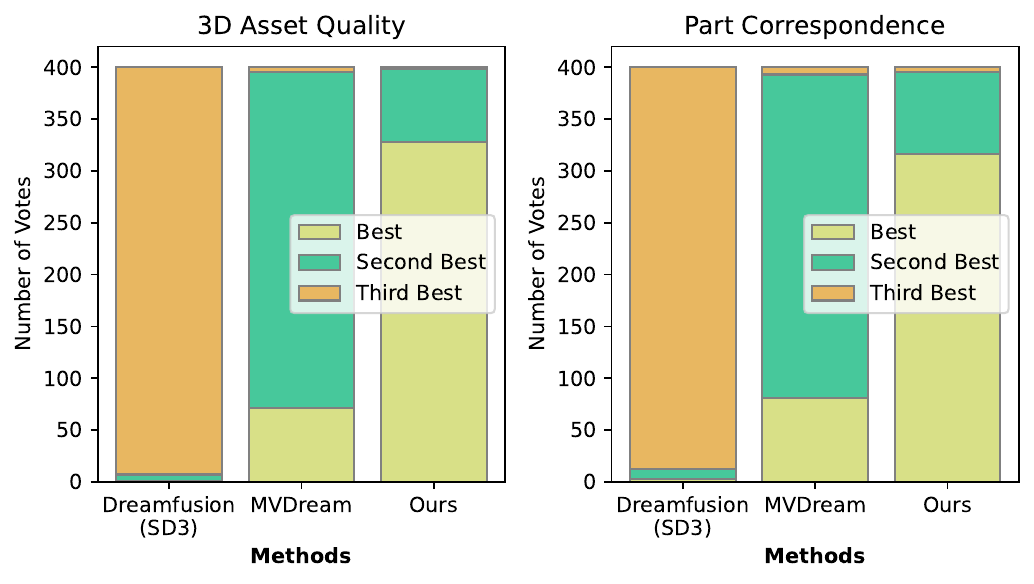}
    \caption{\textbf{Human study results in more detail.} }
    \label{fig:human_study_detail}
\end{figure*}

We present more detailed results statistics in Figure~\ref{fig:human_study_detail}, showing the trend that human evaluators consistently prefer the results generated by \mname.

\section{Non-animal Part-aware Asset Generation}
\label{as:non-animal}
While our main manuscript focuses on generating 3D fantastical beasts, we also observed that our model performs exceptionally well with non-animal, part-specific 3D assets. As demonstrated in Figures~\ref{fig:non_animal_demo_0}, \ref{fig:non_animal_demo_1}, and \ref{fig:non_animal_demo_2}, \mname~continues to excel in generating part-aware 3D non-animal assets, whereas MVDream~\cite{mvdream} struggles with this task. We hope our framework can be extended to more general applications, which we leave for future exploration.

\section{Detailed Running Speed Breakdown}
\label{as:speed}
The Part-Affinity map extraction process takes 41.84 seconds for a single view. The Part-Affinity NeRF requires just 0.06 seconds per optimization step, and the attention-modulated SDS process takes 0.27 seconds per step. Although the part-Affinity map extraction is time-consuming, it is still more efficient and significantly faster than directly applying SDS + SD3.

\begin{table}[]
    \centering
    \tablestyle{1.6pt}{1.0}
    \scalebox{0.85}{
    \begin{tabular}{llccc}
        \toprule
        \multirow{2}{*}{$\alpha_{self}$}& \multirow{2}{*}{$\alpha_{cross}$} & \multicolumn{3}{c}{CLIP Score$\uparrow$} \\
        \cmidrule{3-5}
         & &B/32 & B/16 & L/14 \\
        \midrule
        \multirow{4}{*}{0.8}&0.6&  0.286$\pm 2.3e^{-4}$& 0.288$\pm 3.7e^{-4}$&  0.246$\pm 3.9e^{-4}$\\
        &0.9&  0.285$\pm 2.6e^{-4}$& 0.289$\pm 3.5e^{-4}$& 0.245$\pm 4.6e^{-4}$\\
        &1.2&  0.284$\pm 3.2e^{-4}$&  0.288$\pm 4.5e^{-4}$&   0.242$\pm 5.9e^{-4}$\\
          \midrule
        \multirow{4}{*}{1.2}&0.6& 0.284$\pm 4.2e^{-4}$&  0.289$\pm 3.7e^{-4}$&  0.244$\pm 5.3e^{-4}$  \\
        &0.9&  0.283$\pm 2.7e^{-4}$& 0.288$\pm 4.7e^{-4}$& 0.243$\pm 4.6e^{-4}$\\
        &1.2&  0.284$\pm 2.6e^{-4}$&  0.288$\pm 3.8e^{-4}$&   0.244$\pm 4.4e^{-4}$\\

        \bottomrule
    \end{tabular}}
    \caption{Performance comparison of different attention modulation hyper-parameters. } 
    \label{tab:ab_cross_factor}
\end{table}

\section{Learned Part-Affinity NeRF}
We include 4 videos of the rendered Part-Affinity NeRF on our project website

\section{More Qualitative Results}
\label{as:qualitative}
\begin{figure*}[h!]
    \centering
    \includegraphics[width=0.7\textwidth]{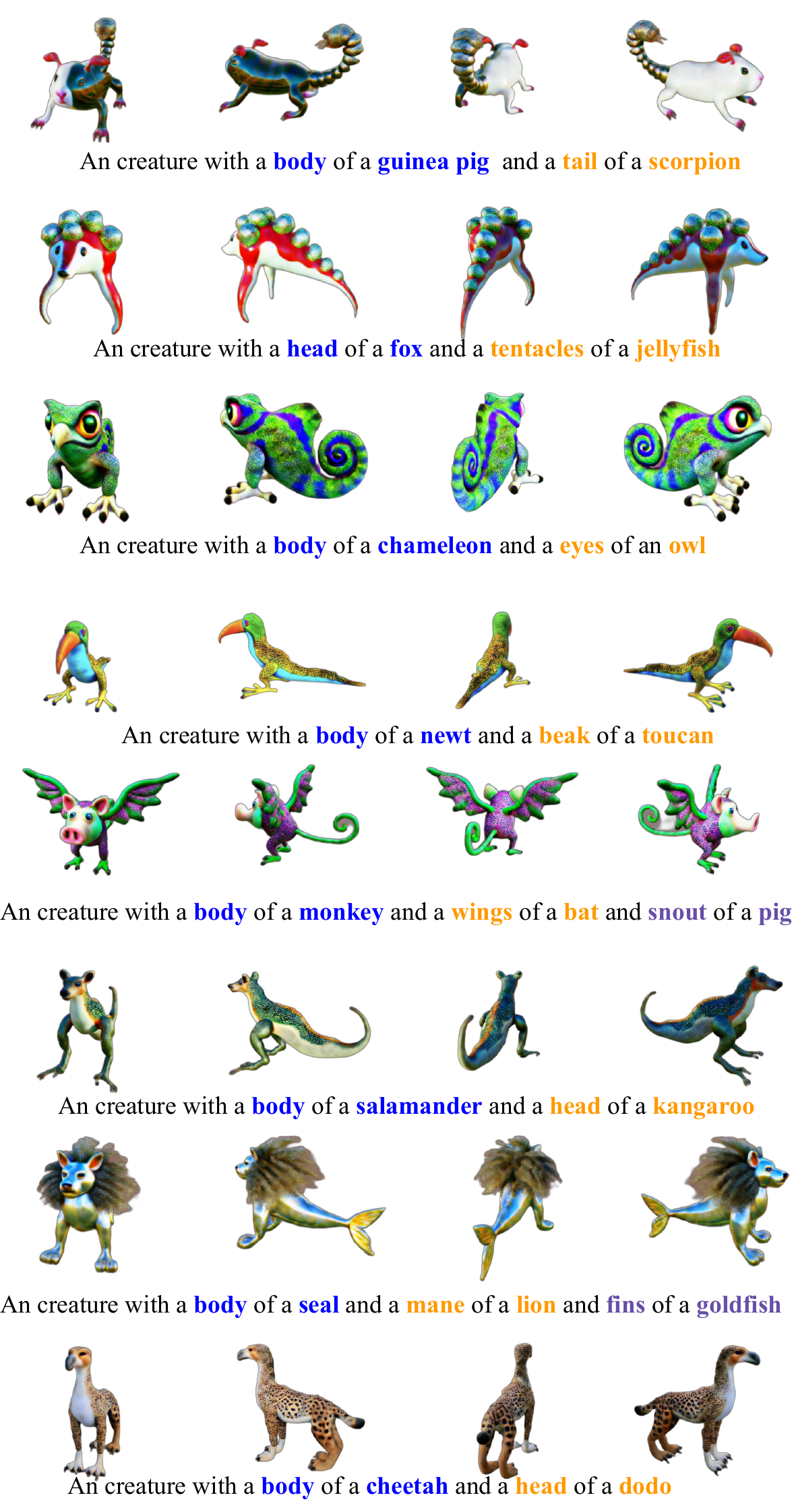}
    \caption{\textbf{More demo results generated from \mname} 
}
    \label{fig:more_demo}
\end{figure*}

We demonstrate more qualitative results in Figure~\ref{fig:more_demo}. All the results show that \mname~can generate part-aware 3D animal assets. We also include 8 rendered videos of RGB, normal maps, and opacity maps on our project website.

\section{Failure Case Analysis}
\label{as:failure}
\begin{figure*}[t!]
    \centering
    \includegraphics[width=0.8\textwidth]{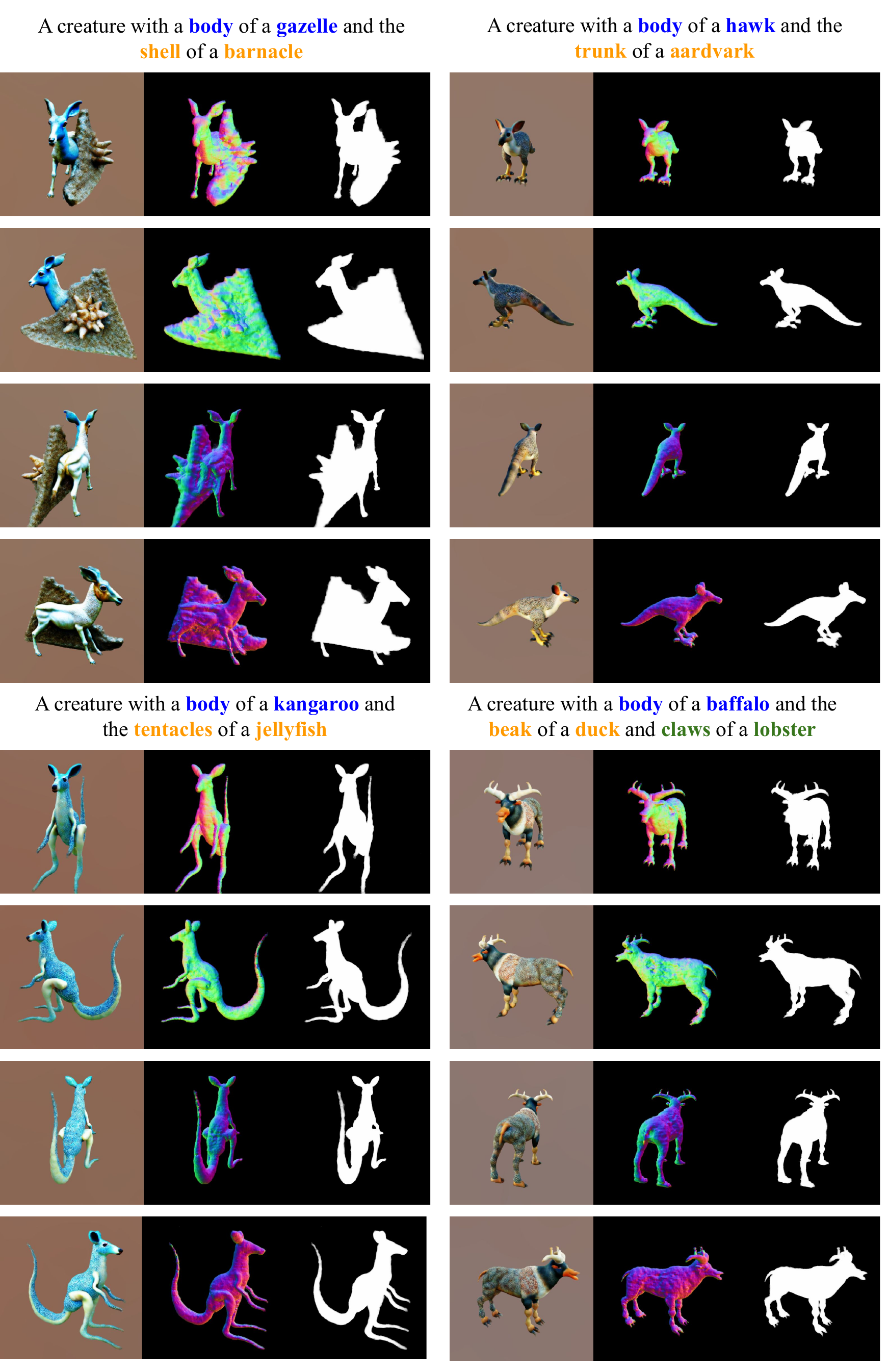}
    \caption{\textbf{Failure case generated by \mname} 
}
    \label{fig:failure}
\end{figure*}

There are also instances where \mname~fails to produce the expected results. The first type of failure occurs when the Part-Affinity map is misplaced, causing the body parts to be incorrectly positioned (as shown in the upper left of Figure~\ref{fig:failure}). The second type of failure happens when two body parts are too similar, making it difficult for \mname~to distinguish between them. For instance, in the upper-right example of Figure~\ref{fig:failure}, the terms “body” and “trunk” are similar, leading to the generated result having a mix of animal features. Additionally, the model sometimes misinterprets parts semantically, as seen in Figure~\ref{fig:non_animal_demo_3}, where “white gun” in the prompt is generated as a “black gun.”


\section{More Implementation Details}
\label{as:implementation}
We chose GPT-4o-mini as the Large Language Model (LLM) to extract part-specific prompts from the original global prompt. 
Additionally, we implemented a part-specific prompt checker to verify that the tokens of the extracted part-specific prompts are also tokens of the original global prompt to prevent hallucinated body part prompts.
However, users also have the option to input part-specific prompts manually, making GPT-4o-mini an optional component in our pipeline.
For the part affinity map extractor, we employed Stable Diffusion 3 medium~\cite{sd3}. 
This cross-attention map operates in the latent space at a resolution of $H=128, W=128$.

The Part-Affinity NeRF is represented by an MLP with a hidden layer of 64 neurons, and we process 128 samples per ray during rendering.  
We use Google Forms to conduct our human studies.

\end{document}